\documentclass[11pt]{article}
\usepackage{coling2020}
\usepackage{times}
\usepackage{url}
\usepackage{latexsym}
\usepackage{graphicx}
\colingfinalcopy

\title{WSL-DS: Weakly Supervised Learning with Distant Supervision for Query Focused Multi-Document Abstractive Summarization}

\author{Md Tahmid Rahman Laskar\textsuperscript{1, 3}, Enamul Hoque\textsuperscript{2}, Jimmy Xiangji Huang\textsuperscript{2, 3}
  \\
 \textsuperscript{1} Department of Electrical Engineering and Computer Science, York University
  \\ 
 \textsuperscript{2}  School of Information Technology, York University
  \\ 
 \textsuperscript{3}  Information Retrieval and Knowledge Management Research Lab, York University
  \\ 
  Toronto, Ontario, Canada 
  \\

  {tahmedge@cse.yorku.ca, enamulh@yorku.ca, jhuang@yorku.ca} \\

  }

\date{}

\begin{document}
\maketitle

\begin{abstract}
In the Query Focused Multi-Document Summarization (QF-MDS) task, a set of documents and a query are given where the goal is to generate a summary from these documents based on the given query. However, one major challenge for this task is the lack of availability of labeled training datasets. To overcome this issue, in this paper, we propose a novel weakly supervised learning approach via utilizing distant supervision. In particular, we use datasets similar to the target dataset as the training data where we leverage pre-trained sentence similarity models to generate the weak reference summary of each individual document in a document set from the multi-document gold reference summaries. Then, we iteratively train our summarization model on each single-document to alleviate the computational complexity issue that occurs while training neural summarization models in multiple documents (i.e., long sequences) at once. Experimental results in Document Understanding Conferences\footnote{https://duc.nist.gov/} (DUC) datasets show that our proposed approach sets a new state-of-the-art result in terms of various evaluation metrics.

\end{abstract}

\section{Introduction}
With the rapid growth of textual documents on the internet, accessing information from the web has become a challenging issue \cite{yao2017survey}. Often users want the summary of a topic from various sources to fulfill their information needs \cite{queryfocusedsummarization2017unsupervised}. The QF-MDS task deals with such problems where the goal is to summarize a set of documents to answer a given query.

In the QF-MDS task, the summaries generated by the summarizer can be either extractive or abstractive~\cite{yao2017survey,kulkarni2020aquamusegoogle}. An extractive summarizer extracts relevant text spans from the source document(s), whereas an abstractive summarizer generates a summary in natural language that may contain some words which did not appear in the source document(s) \cite{rush,nallapati2016abstractive,nema}. With the rising popularity of virtual assistants in recent years, there is a growing interest to integrate abstractive summarization capabilities in these systems for natural response generation \cite{nishida2019multi}.

One major challenge for the QF-MDS task is that the datasets (e.g., DUC 2005, 2006, 2007) used for such tasks do not contain any labeled training data. Therefore, neural summarization models that leverage supervised training cannot be used in these datasets. Note that for other related tasks \cite{allan2003challenges,liu2008modeling,miao2012proximity}, how to reduce the demands for labeling the data and how to leverage unlabeled data were also identified as a major challenge. While using datasets similar to the target dataset as the training data for the QF-MDS task, we find that these datasets only contain multi-document gold summaries. However, the state-of-the-art transformer-based \cite{vaswani2017attention} summarization models \cite{liuemnlpbertsum,laskar2020query} cannot be used in long documents due to computational complexities \cite{beltagy2020longformer,zaheer2020bigbird}. To tackle these issues, we propose a novel weakly supervised approach by utilizing distant supervision to generate weak reference summary of each single-document from multi-document gold reference summaries. We train our model on each document with weak supervision and find that our proposed approach that generates abstractive summaries is very effective for the QF-MDS task. More concretely, we make the following contributions: 

\begin{itemize}

\item First, to address the issue of unlabeled individual documents in a training document set, we utilize pre-trained sentence similarity models \cite{liu2019roberta,laskar-LREC} to generate the weak reference summary of each individual document from multi-document gold reference summaries.

\item Second, to address the computational issue to train neural models in long documents \cite{zaheer2020bigbird,beltagy2020longformer}, we propose an iterative approach that adopts a pre-trained single-document generic summarization model to leverage the effectiveness of fine-tuning such models for query focused abstractive summarization \cite{laskar2020query} and extends it for the QF-MDS task.

\item Experimental results on DUC 2005-07 datasets show that our proposed approach sets a new state-of-the-art result in terms of various ROUGE scores. As a secondary contribution, we will make our source codes publicly available here: \url{https://github.com/tahmedge/WSL-DS-COLING-2020}. 
\end{itemize}

\section{Related Work}
Early work on multi-document summarization was mostly focused on generic summarization \cite{nayeem2018abstractive}, whereas the amount of work for QF-MDS had been very limited \cite{yao2017survey}. Due to the lack of training data for the QF-MDS task, most previous works were based on various unsupervised approaches that could only generate extractive summaries \cite{bi-plsa,hiersum,multimr,spopt,qode,ctsum,queryfocusedsummarization2016unsupervised,queryfocusedsummarization2017unsupervised}.

To generate the abstractive summaries for the QF-MDS task, \cite{baumel2018query} proposed a transfer learning technique to tackle the issue of no training data. They adopted the Pointer Generation Network (PGN) \cite{see} pre-trained for the generic abstractive summarization task in a large dataset to predict the query focused summaries in the target dataset via modifying the attention mechanism of the PGN model. However, their model failed to outperform different extractive approaches in terms of various ROUGE scores \cite{queryfocusedsummarization2017unsupervised,dualces}. 

Identifying sentences which are relevant to the query is an important step for the QF-MDS task. For this purpose, various approaches were utilized such as counting word overlaps \cite{baumel2018query} or the Cross-Entropy Method \cite{queryfocusedsummarization2017unsupervised}. Though neural models based on supervised training have significantly outperformed various non-neural models for the answer selection task in recent years \cite{laskarbertammcs,laskar-LREC}, such neural models have not been effectively used for the QF-MDS task yet due to the absence of labeled data for the relevant sentences in the QF-MDS datasets. 

Recently, \cite{tanda2019} showed that neural models pre-trained in a large Question Answering (QA) dataset could effectively select answers in other QA datasets. More recently, such pre-trained answer selection models for the QF-MDS task were used by \cite{lapata2020query}. In their work, they utilized distant supervision from various QA datasets using the fine-tuned BERT \cite{devlin2018bert} model to filter out the irrelevant sentences from the documents. However, \cite{baumel2018query} showed that filtering sentences as an early step could lead to performance deterioration for the QF-MDS task. 
Thus, instead of applying distant supervision to filter out some sentences from the document, we apply it to generate the weak reference summary of each unlabeled document in our training datasets. Our proposed weakly supervised learning approach not only allows us to leverage the advantage of fine-tuning pre-trained generic summarization models \cite{laskar2020query}, but also allows us to overcome the limitation of training neural models in long documents \cite{beltagy2020longformer,zaheer2020bigbird}.  

\section{Our Proposed Approach}

\begin{figure*}[t!]
\begin{center}
\includegraphics[width=\linewidth]{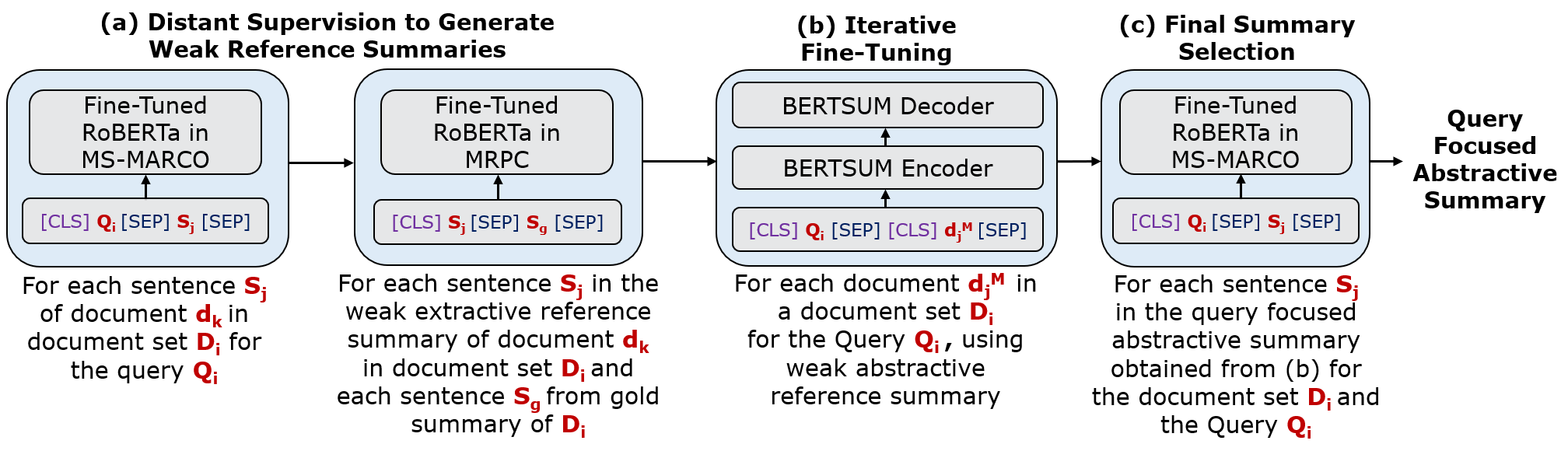}
\caption{
An overview of our model that generates (a) weak reference summary using \textbf{RoBERTa} for (b) iterative fine-tuning using \textbf{BERTSUM} to generate query focused abstractive summaries which are then ranked by (c) \textbf{RoBERTa}. [CLS] and [SEP] are the special tokens used with inputs \cite{devlin2018bert}.
}

\label{fig:ModelOverview}
\end{center}

\end{figure*}
Suppose, we have a query $Q = q_1, q_2,...,q_k$ containing $k$ words and a set of $N$ documents $D$ = $d_1, d_2, ...,d_N$. For the QF-MDS task, the goal is to generate a summary $S = s_1, s_2,...s_n$ containing $n$ words from the document set $D$ for the given query $Q$.

In figure \ref{fig:ModelOverview}, we show the overall architecture of our proposed approach. Since there is no training data available for the QF-MDS task, we provide supervised training to our target dataset by using other QF-MDS datasets as the training data. However, the available QF-MDS datasets \cite{queryfocusedsummarization2017unsupervised} only contain the gold summaries generated by human experts from multiple documents and do not contain the gold summary of each individual document. Due to the limitations of using neural models in long documents \cite{beltagy2020longformer,zaheer2020bigbird}, we propose an iterative approach to train our model on each document of a document set. For this purpose, we generate the weak reference summary of each document from the multi-document gold summaries using distant supervision to train our model for the QF-MDS task. Finally, we rank the generated query focused summaries via an answer selection model \cite{laskar-LREC}. In the following, we give a detailed description of our proposed approach. 

\subsection{Weakly Supervised Learning with Distant Supervision}

To generate the weak reference summaries using distant supervision, we utilize the pre-trained RoBERTa model \cite{liu2019roberta} in two steps (see Figure \ref{fig:ModelOverview}a). At first, we generate the weak extractive reference summary of each individual document using a RoBERTa sentence similarity model fine-tuned for the \textit{answer selection} task. Then, we measure the similarity score between each sentence in the human written (abstractive) multi-document gold summaries with each sentence in the weak extractive reference summary using a RoBERTa sentence similarity model fine-tuned for the \textit{paraphrase identification} task. Based on the similarity score, we select the most relevant sentences from the gold reference summaries as the weak abstractive reference summary for each document. Below we describe these steps in detail. 

\paragraph{RoBERTa Answer Selection Model:}
In this step, we first generate the initial weak extractive reference summary of each individual document $d_k$ by measuring the relevance between the query $Q_i$ and each sentence $S_j$ in $d_k$.
For this purpose, we adopt the RoBERTa sentence similarity model from \cite{laskar-LREC} for its impressive performance in the answer sentence selection task and fine-tune it in the QA-ALL dataset of MS-MARCO \cite{msmarco}. The fine-tuned RoBERTa\textsubscript{MS-MARCO} model was then utilized in our training dataset to measure the similarity score between each sentence in the document and the query. Based on the similarity score, we select the top $K=3$ most relevant sentences as the weak extractive reference summary. Note that we use the value of $K=3$ because extracting only $3$ sentences was found effective in different extractive summarizers such as the LEAD-3 baseline \cite{see,liuemnlpbertsum}, as well as the BERTSUM\textsubscript{EXT} model \cite{liuemnlpbertsum}. 
\paragraph{RoBERTa Paraphrase Identification Model:} We provide distant supervision to generate the weak abstractive reference summary by replacing each sentence in the weak extractive reference summary (generated in the previous step) with the most similar sentence found in the multi-document gold summaries. For this purpose, we fine-tune the RoBERTa model for the paraphrase identification task in the MRPC dataset \cite{liu2019roberta}. Then for each document $d_k$ in a document set $D_i$, we utilize the fine-tuned RoBERTa\textsubscript{MRPC} paraphrase identification model to replace each sentence $S_j$ in the weak extractive reference summary of $d_k$ with the most similar sentence $S_g$  found in the gold summaries (among the sentences that are not already selected for the same document) of $D_i$.

\subsection{Iterative Fine-Tuning for Multi-Document Summarization}

 For the QF-MDS task, we adopt the transformer-based \cite{vaswani2017attention} BERTSUM model pre-trained for generic abstractive summarization on the CNN/DailyMail dataset \cite{liuemnlpbertsum} to leverage the advantages of fine-tuning it for the query focused abstractive summarization task \cite{laskar2020query}. However, BERTSUM was trained for the single-document summarization task by considering at most 512 tokens \cite{liuemnlpbertsum,beltagy2020longformer,zaheer2020bigbird}. To address this issue for the multi-document scenario, we take an iterative approach (see Figure \ref{fig:ModelOverview}b). At first, we incorporate query relevance via concatenating the query with each document, similar to the work of \cite{laskar2020query}. Then, we fine-tune BERTSUM using the weak  abstractive reference summary to generate the query focused abstractive summary of each document in a document set. The sentences in the generated query focused summaries of each document set are then ranked using the fine-tuned RoBERTa\textsubscript{MS-MARCO} answer selection model to select the  sentences that are most relevant to the query (see Figure \ref{fig:ModelOverview}c).

\section{Experimental Setup}

We now describe the datasets used in this paper, followed by the details of our implementation. 
\paragraph{Datasets:}
We use the DUC 2005, 2006, and 2007 datasets for the QF-MDS task.
The number of document sets were 50, 50, and 45 where  the number of documents in each document set were 32, 25, and 25 in DUC 2005, 2006 and 2007 datasets respectively~\cite{queryfocusedsummarization2017unsupervised}. Each document set is associated with a query and the objective is to generate a summary containing at most 250 words from the document set based on the given query. Given the absence of the training data, to evaluate our model in each year's dataset we use the datasets from the other two years for training. From each year's training data, we randomly selected 20\% of the document sets for validation while we used the rest for training.

\paragraph{Implementation:}
For the RoBERTa model, we used its Large version \cite{liu2019roberta,laskar-LREC} and implemented using HuggingFace's Transformer \cite{wolf2019huggingface}. For fine-tuning the summarization model, we used the BERTSUM\textsubscript{EXT-ABS}\footnote{https://github.com/nlpyang/PreSumm} model pre-trained on the CNN/DailyMail dataset \cite{liuemnlpbertsum}. While selecting the most relevant sentences as the final query focused summary, we used the Trigram Blocking to reduce redundancy \cite{paulus2018deepICLR}.
To fine-tune the BERTSUM model, we kept most parameters similar to the original work \cite{liuemnlpbertsum} and ran 50 steps with batch size equal to 200. Among these 50 steps, we selected the step for evaluation that performed the best on the validation set. All of our models were run in multi-GPU settings using 4 NVIDIA V100 GPUs. We report the results based on both Recall and F1 scores in terms of ROUGE-1, ROUGE-2, and ROUGE-SU4 metrics \cite{rouge}. 
From now on, we denote ROUGE as \textbf{R}.

\begin{table*}[t!]
\centering
\small
\begin{center}
\begin{tabular}{c|c|c|c|c|c|c|c|c|c}
\cline{2-10}
\multicolumn{1}{c|}{\textbf{(a) F1 Score:}} &
\multicolumn{3}{c|}{\textbf{DUC 2005}} & 
\multicolumn{3}{c|}{\textbf{DUC 2006}} & 
\multicolumn{3}{c}{\textbf{DUC 2007}} \\ \cline{1-10}
\centering
\textbf{Model} & \textbf{\textbf{R-1}} & \textbf{R-2} & \textbf{\textbf{R-SU4}} 
& \textbf{\textbf{R-1}} & \textbf{R-2} & \textbf{\textbf{R-SU4}} 
& \textbf{\textbf{R-1}} & \textbf{R-2} & \textbf{\textbf{R-SU4}} 
\\ \hline
\textbf{\cite{queryfocusedsummarization2017unsupervised} *} & 37.78 & 7.45 & 13.02 & 40.47 & 9.13 & 14.73 & 42.86 & 11.34 & 16.53 
\\ 
\textbf{\cite{lapata2020query} *} & - & - & - & 41.6 & 9.5 & 15.3 & 43.3 & 11.6 & 16.8
\\ 
\textbf{\cite{dualces} *} & 38.08 & 7.54 & 13.17 & 41.23 & 9.47 & 14.97 & 43.24 & 11.78 & 16.83
\\ 
\hline
\textbf{PQSUM\textsubscript{EXT} *} & 37.52 & 7.84 & 13.29 & 40.68 & 9.29 & 14.66 & 42.57 & 11.20 & 15.98
\\ 
\textbf{PQSUM\textsubscript{ABS}} & 38.35 & 7.94 & 13.44 & 40.87 & 9.43 & 14.83 & 42.17 & 10.82 & 15.98
\\
\hline
\textbf{PQSUM\textsubscript{WSL-DS}} & \textbf{40.32} & \textbf{9.17} & \textbf{14.73} & \textbf{43.49} & \textbf{10.78} & \textbf{16.45} & \textbf{44.72} & \textbf{12.44} & \textbf{17.72} 
\\ 
\hline
 \multicolumn{10}{c}{} \\
\cline{2-10}
\multicolumn{1}{c|}{\textbf{(b) Recall:}} &
\multicolumn{3}{c|}{\textbf{DUC 2005}} & 
\multicolumn{3}{c|}{\textbf{DUC 2006}} & 
\multicolumn{3}{c}{\textbf{DUC 2007}} \\ \cline{1-10}
\textbf{Model} & \textbf{\textbf{R-1}} & \textbf{R-2} & \textbf{\textbf{R-SU4}} 
& \textbf{\textbf{R-1}} & \textbf{R-2} & \textbf{\textbf{R-SU4}} 
& \textbf{\textbf{R-1}} & \textbf{R-2} & \textbf{\textbf{R-SU4}} 
\\ \hline
\textbf{\cite{queryfocusedsummarization2017unsupervised} *} & 40.35 & 7.94 & 13.91 & 43.01 & 9.69 & 15.65 & 45.45 & 12.02 & 17.54
\\ 
\textbf{\cite{baumel2018query}} & 39.82 & 6.98 & \textbf{15.73} & 42.89 & 8.73 & \textbf{17.75} & 43.92 & 10.13 & \textbf{18.54}
\\ 
\textbf{\cite{dualces} *} & \textbf{40.82} & 8.07 & 14.13 & \textbf{43.94} & 10.09 & 15.96 & \textbf{46.02} & \textbf{12.53} & 17.91
\\ 
\hline
\textbf{PQSUM\textsubscript{EXT} *} & 37.55 & 7.84  &  13.31  &  40.41  &  9.22  & 14.56 &  42.41 &  11.08 &  15.92
\\ 
\textbf{PQSUM\textsubscript{ABS}} & 38.36 & 7.92 & 13.43 & 40.59 & 9.39 & 14.73 & 42.05 & 10.79 & 15.91 
\\ 
\hline
\textbf{PQSUM\textsubscript{WSL-DS}} & 40.36 & \textbf{9.17} & 14.74 & 43.22 & \textbf{10.70} & 16.35 & 44.61 & 12.40 & 17.66
\\ 
\hline
\end{tabular}
\caption{Performance comparisons in terms of (a) \textbf{F1} and (b) \textbf{Recall}. `*' denotes extractive model.}
  \label{tab:all} 
\end{center}
\end{table*}
\begin{table*}[t!]
\small
  \centering
  \label{tab:freq}
\renewcommand{\arraystretch}{1} 
  \begin{tabular}{l|l|l|l}
    \hline
    \textbf{Model}&\textbf{Recall}&\textbf{F1}&\textbf{Statistically Significant}\\
  \hline
    \textbf{PQSUM\textsubscript{WSL-DS}} & \textbf{42.73} & \textbf{42.84} &  \\
        \textbf{without Distant Supervision} & 41.77 \tiny{(- 2.25\%)} & 41.88 \tiny{(- 2.24\%)} & \textbf{No}, based on paired t-test ($p$ $\leq$ $.05$) \\
        \textbf{without Trigram Blocking} & 40.92 \tiny{(- 4.24\%)} & 41.01 \tiny{(- 4.27\%)} & \textbf{No}, based on paired t-test ($p$ $\leq$ $.05$) \\
         \textbf{without Weakly Supervised Learning} & 40.01 \tiny{(- 6.37\%)} & 40.12 \tiny{(- 6.35\%)} & \textbf{Yes}, based on paired t-test ($p$ $\leq$ $.05$) \\
 \hline
  \end{tabular}
\caption{Ablation test result based on the average \textbf{R-1}. `-' denotes `deterioration from original model'.}
  \label{tab:ab}
\end{table*}
\section{Results and Discussions}
We now analyze the performance of our proposed model by comparing with other models (see Table \ref{tab:all}). We also perform ablation test to  further investigate its effectiveness (see Table \ref{tab:ab}). We denote our approach of using the \textbf{P}re-trained models (RoBERTa and BERTSUM) for \textbf{Q}uery focused \textbf{SUM}mary generation via utilizing Weakly Supervised Learning with Distant Supervision (WSL-DS) as \textbf{PQSUM\textsubscript{WSL-DS}}. For performance comparisons, we use two baselines that do not utilize weak supervision and fine-tuning.  Note that both of these baselines use the BERTSUM \cite{liuemnlpbertsum} model pre-trained on the CNN/DailyMail dataset. One of them is pre-trained for extractive summarization: \textbf{PQSUM\textsubscript{EXT}}; while the other is pre-trained for abstractive summarization: \textbf{PQSUM\textsubscript{ABS}}. These baselines generate the summaries of all documents in a document set which are then ranked using  RoBERTa\textsubscript{MS-MARCO}. Moreover, we compare our model with four recent works: i) CES-50 \cite{queryfocusedsummarization2017unsupervised}, ii) RSA \cite{baumel2018query}, iii) QUERYSUM \cite{lapata2020query}, and iv) DUAL-CES \cite{dualces}.

\paragraph{Performance Comparisons:}
From Table \ref{tab:all}(a), we find that our \textbf{PQSUM\textsubscript{WSL-DS}} model sets a new state-of-the-art in all datasets based on the F1 metric for all ROUGE scores.
Specifically, based on \textbf{R-1}, it improves by 5.88\% in DUC 2005 from \cite{dualces} along with 4.54\% and 3.28\% from \cite{lapata2020query} in DUC 2006 and 2007 respectively.
From Table \ref{tab:all}(b), we find that our model also sets a new state-of-the-art in terms of \textbf{R-2} Recall in DUC 2005 and 2006, but fails to outperform the  DUAL-CES \cite{dualces} in DUC 2007. 
In comparison to the abstractive RSA model \cite{baumel2018query}, we find that our model outperforms them in all datasets in terms of both \textbf{R-1} and \textbf{R-2} Recall, but fails to outperform them in \textbf{R-SU4} scores.
Moreover, we find based on paired t-test ($p$ $\leq$ $.05$) that the weakly supervised learning \textbf{significantly} outperforms the  baselines in terms of both Recall and F1.

\paragraph{Ablation Test:}

The result of our ablation test based on the average of \textbf{R-1} scores across all datasets is shown in Table \ref{tab:ab}. We find that the performance is deteriorated if we exclude \textit{Distant Supervision} via removing the RoBERTa\textsubscript{MRPC} model, as well as if the \textit{Trigram Blocking} is not used. Moreover, the performance is \textbf{significantly} degraded if the summary is generated by only ranking the sentences in the documents using the fine-tuned RoBERTa\textsubscript{MS-MARCO} without utilizing \textit{Weakly Supervised Learning}. 
\section{Conclusions and Future Work}

In this paper, we propose a novel weakly supervised approach for the Query Focused Multi-Document Abstractive Summarization task to 
tackle the issue of no available labeled training data for such tasks. We also propose an iterative approach to address the computational problem that occurs while training neural models in long documents \cite{2020reformer,beltagy2020longformer,zaheer2020bigbird}.
Experimental results in three datasets show that our proposed approach sets a new state-of-the-art result in various evaluation metrics. In the future, we will apply our models on more tasks, such as information retrieval applications \cite{JH1,JH2,JH3,Jh4}, sentiment analysis \cite{JH5,JH6}, learning from imbalanced or unlabeled datasets \cite{JH7,bari2019zero,bari2020multimix}, and automatic chart question answering \cite{kim2020answering}. 

\section*{Acknowledgements}
We gratefully appreciate the area chair(s) and the reviewers for their excellent review comments. This research is supported by the Natural Sciences \& Engineering Research Council (NSERC) of Canada, the York Research Chairs (YRC) program and an ORF-RE (Ontario Research Fund-Research Excellence) award in BRAIN Alliance. We acknowledge Compute Canada for providing us the computing resources.

\bibliographystyle{coling}
\bibliography{coling2020}

\end{document}